\documentclass[conference]{IEEEtran}
\IEEEoverridecommandlockouts
\usepackage{amsmath,amssymb,amsfonts}
\DeclareMathOperator{\SSIM}{SSIM}
\usepackage{graphicx}
\usepackage{textcomp}
\usepackage{chngpage}
\usepackage{algorithm2e}
\SetKwComment{Comment}{/* }{ */}
\usepackage{calc}
\usepackage{algpseudocode}
\usepackage{xcolor}
\usepackage{mathtools,array}
\DeclarePairedDelimiter{\abs}{\lvert}{\rvert}
\graphicspath{ {images/} }
\def\BibTeX{{\rm B\kern-.05em{\sc i\kern-.025em b}\kern-.08em
    T\kern-.1667em\lower.7ex\hbox{E}\kern-.125emX}}
\begin{document}

\title{Localized Super Resolution for Foreground Images using U-Net and MR-CNN\\}

\author{\IEEEauthorblockN{Umashankar Kumaravelan}
\IEEEauthorblockA{
umashanks99@gmail.com\\Vellore Institude of Technology, Chennai, India}
\and
\IEEEauthorblockN{Nivedita M}
\IEEEauthorblockA{
nivedita.m@vit.ac.in\\Vellore Institude of Technology, Chennai, India}
}

\maketitle

\begin{abstract}
 Images play a vital role in understanding data through visual representation. It gives a clear representation of the object in context. But if this image is not clear it might not be of much use. Thus, the topic of Image Super Resolution arose and many researchers have been working towards applying Computer Vision and Deep Learning Techniques to increase the quality of images. One of the applications of Super Resolution is to increase the quality of Portrait Images. Portrait Images are images which mainly focus on capturing the essence of the main object in the frame, where the object in context is highlighted whereas the background is occluded.  When performing Super Resolution the model tries to increase the overall resolution of the image. But in portrait images the foreground resolution is more important than that of the background. In this paper, the performance of a Convolutional Neural Network (CNN) architecture known as U-Net for Super Resolution combined with Mask Region Based CNN (MR-CNN) for foreground super resolution is analysed. This analysis is carried out based on Localized Super Resolution i.e We pass the LR Images to a pre-trained Image Segmentation model (MR-CNN) and perform super resolution inference on the foreground or Segmented Images and compute the Structural Similarity Index (SSIM) and  Peak Signal-to-Noise Ratio (PSNR) metrics for comparisons.
\end{abstract}


\section{Introduction}
The challenging task of computing a high resolution image from a single low resolution has given rise to multiple state-of-the-art architectures and many researchers are still working towards new methods and architectures to increase the performance of previous models. Many of the papers aim at improving or modifying the model architecture or the loss and optimizer functions to increase performance. The influence of the image region also plays a vital role in super resolution task, since the main goal is to increase resolution such that the object or content in context is more clearly visible. Images such as portrait images, passport photos, etc. all have a common characteristic i.e the foreground is clear whereas the background is blurred or occluded. For such images quality is a key aspect and moreover, it is the foreground resolution that matters. In this work we propose a combination of Image Segmentation Models for the task of Foreground Super Resolution. The proposed model is a U-Net based CNN architecture with MR-CNN for computing the Super Resolution output of the foreground or segmented Image. The output of MR-CNN model i.e. the segmented super resolution output is added back to the LR image and then the evaluation metrics such as SSIM and PSNR between the Segmented Super Resolution (SR) Output and normal SR output are computed to evaluate the increase in metric performance.

\section{Literature Survey}
The first deep learning method \cite{SRCNN} of leveraging the use of Convolutional Neural Networks \cite{cnn} (CNN) and achieve state-of-the-art results is titled "Image Super-Resolution Using Deep Convolutional Networks" or commonly known as SRCNN. This architecture consists of three layers i.e. One layer for patch extraction which is then passed into the non-linear mapping layer which finally is given to the reconstruction layer. The patch layer extracts dense patches from the image. The non-linear layer consists of 1x1 convolutional filters which change or increase the number of channels and includes a non-linearity factor to it. And finally the reconstruction layer maps the output from the non-linear layer to a high resolution representation. This gave rise to multiple new architectures such as Very Deep Super Resolution (VDSR) \cite{vdsr} which is 20 layers deeper than SRCNN, Fast Super-Resolution Convolutional Neural Network (FSRCNN) which includes convolutional layers for shrinking, non-linear mappping and expansion \cite{fsrcnn} and the Effiecient Sub Pixel CNN (ESPCN) \cite{espcn} implements sub-pixel shuffle layers after which upscaling is performed. Similarily, the introduction of a new type of CNN Architecture known as Residual CNNs or ResNets \cite{resnet} paved way for residual based Super Resolution Architectures such as Enhanced Deep Residual Networks for Single Image Super-Resolution (EDSR) which performs optimization by removing unnecessary residual layers and perform optimized expansion of the model \cite{edsr} and Wide Activation for Efficient and Accurate Image Super-Resolution (WDSR) which implements a low-rank convolution to widen activation. \cite{wdsr}. 
\\
\\
Following the lines of CNN Architectures, we shall take a look at a particular CNN Architecture known as the U-Net Architecture \cite{unet}. This architecture can be understood by splitting it into to individual paths, the downsampling path and upsampling path. The input passes through a series of convolutional layers and downsized to a suitable representation. This feature map is then passed through a series of upsampling layers to generate an output with the original representation shape. The convolutional layers in the downsampling and upsampling path have skip connections connecting each downsample layer to the corresponding upsample layer where the information is concatenated. This is done to provide transfer information representation from downsampling layers to the upsampling layers. This architecture is mainly used for the purpose of image segmentation, where the model is trained with images as input and corresponding object segmentation maps as output. The original paper was proposed for image segmentation in biomedical-imaging. In this paper, we have leveraged the U-Net architecture for the purpose of Super Resolution to avoid usage of intensive compute architectures by using large pretrained models and still generate reasonable SR images.
\\
\\
 The task of object detection and image segmentation has been gaining attention ever since the rise of CNN Architectures. Many computer vision tasks and applications first need to detect, classify and segment the objects from the image so that the necessary image processing tasks can be carried out. One such famous and state-of-the-art architecture is known as MRCNN \cite{mrcnn} or Mask R-CNN. This architecture is a combination of both object detection and image segmentation models. The first part of the model is known as the Region proposal network or RPN network which proposes the object bounding boxes. The second part is known as the Binary mask classifier which generates the object segmentation masks for every class. The backbone of the model is a ResNet101 Architecture. Our proposed architecture combines both the U-Net architecture with the MR-CNN to generate high resolution foregorund object segmentations.
\\
\\
Foreground/Background Segmentation is the process of segmenting the foreground objects from the background using sgement maps and vice versa. The authors of \cite{vggfor} propose a VGG-16 based CNN for feature extraction and the resulting feature maps are then upsampled using deconvolution layer and then sigmoid thresholding is carried out to get background foreground information. This implies they had leveraged a CNN model to detect foreground objects which is then segmented based on sigmoid map which depicts the difference between object in context pixels and background pixels. Similarily the authors of \cite{scenerem} train a CNN Model on scenic datasets by using only the scene backgrounds as the input. This model segments out the backgrounds alone from an image. This model was trained only with scene-specific knowledge, hence it cannot be used to segment out specific objects from a given image. An object segmentation model is a better choice incase the usecase requires needs specific object segments. Thus we have leveraged the use of MR-CNN for such case.
\\
\\
 Some of the applications of our architecture include increasing quality of foreground images /objects in video conference platforms such as Zoom, Teams, etc. Replacing the background with another background, filter effects and video editing. The authors of \cite{rhbm} propose a real-time high quality background replacement architecture where the current background of any image can be changed keeping the existing foreground objects intact. The results are so clear that they preserve the details of the object while merging/blending them with the background. This architecture achieved 30 FPS in 4k resolution and 60 FPS on HD resolution. Similar to our architecture this paper also implements a segmentation model similar to MRCNN and DeeplabV3 Models. They have implemented the segmentation model with Atrous Spatial Pyramid Pooling with DeepLabV3 \cite{deeplab}. The decoder network which is used to predict the Alpha matte of the background, the foreground mask and the foreground segmented image is made up of bilinear upsampling layers concatenated with bilinear upsampling at each step followed by 3x3 convolution and ReLu activation functions. For each convolutional layer, Batch Normalization is applied. While this paper focuses on effortless blending and merging of foreground with new backgrounds, our architecture focuses on increasing the quality of foreground objects alone while disregarding background during inference.

\section{Proposed Architecture}
\begin{figure}[htbp]
\centerline{\includegraphics[width=8cm, height=8cm]{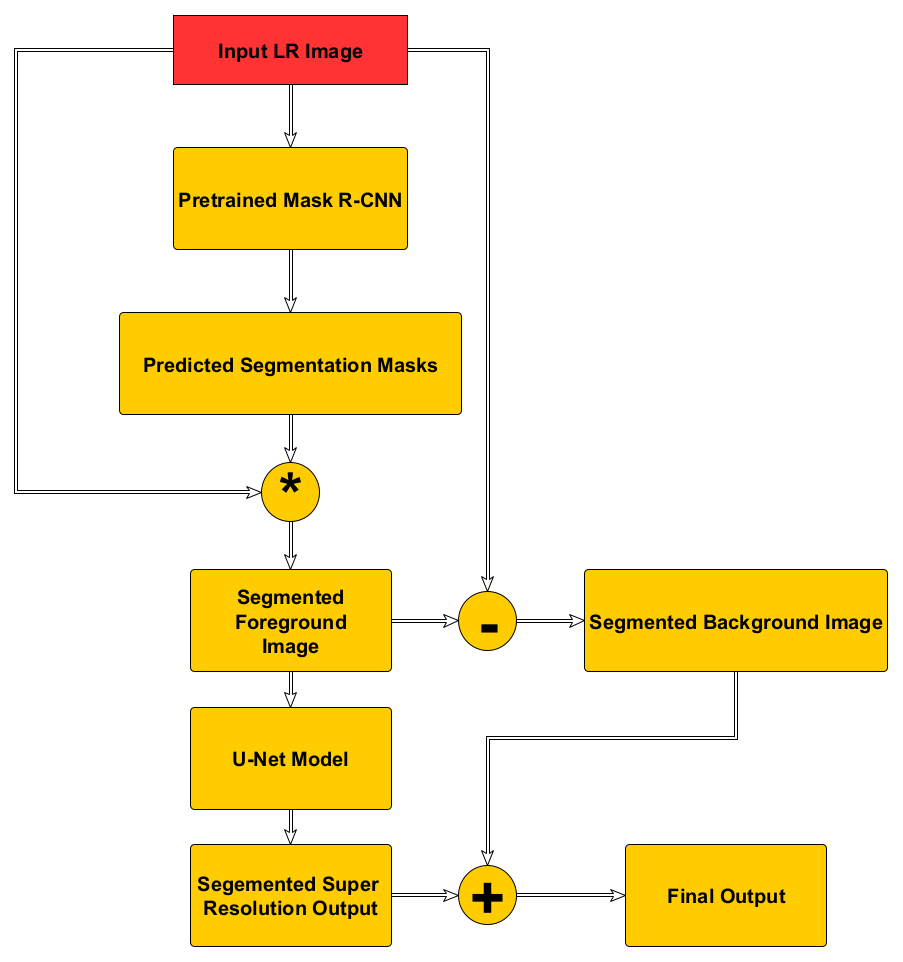}}
\caption{Proposed Architecture Flow for Foreground Super Resolution using U-Net and MR-CNN.}
\label{proparch}
\end{figure}

The U-Net Architecture is a CNN architecture that expanded with few changes in the CNN architecture. It was developed to carry out image segmentation on biomedical images where the target is to segment and highlight the area of infection. We leverage this model for the task of super resolution in our experiments. We modeled a simple Vanilla CNN U-Net and trained the model on 800 LR Images which were scaled down by a factor of 50 from their corresponding HR Images. For Foreground Super Resolution we take a test Image and convert it to the corresponding LR Image. This LR image is then passed into a pretrained MR-CNN to obtain the foreground or object segmentation mask. The mask is then used to extract the foreground alone and this segmented LR image is passed into the trained U-Net to obtain the corresponding super resolution output. The SR Image is then added back to the segmented LR Image to get the final Super Resolution output. The flow of the proposed architecture is shown in Figure \ref{proparch}.

\section{Implementation and Training}
\subsection{Dataset and Augmentation}
\begin{figure}[htbp]
\centerline{\includegraphics[width=5cm, height=5cm]{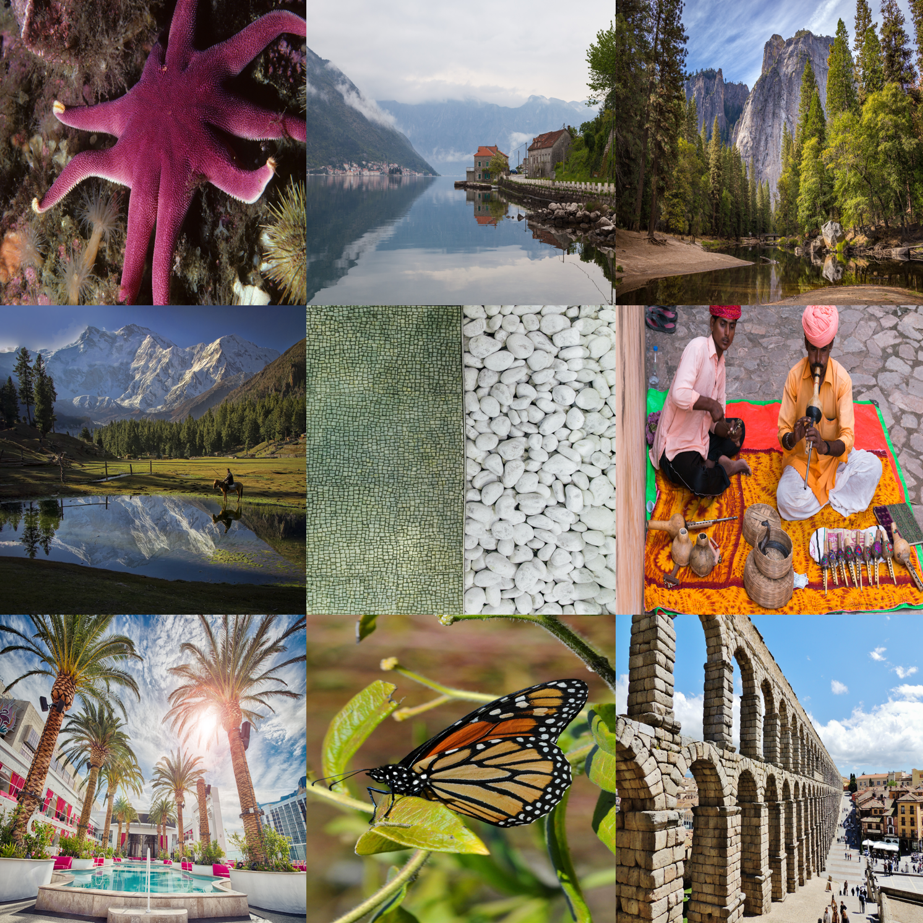}}
\caption{Subset of Images from DIV2K Training Set.}
\label{div2kset}
\end{figure}

The DIV2K \cite{div2k} dataset is collection of high resolution RGB Images which was compiled for the NTIRE2017 and NTIRE2018 Super-Resolution Challenges. DIV2K dataset stands for DIVerse 2K Resolution Images. Each of these 1000 images have atleast 2K pixels on one of their axes (vertical or
horizontal).
The entire dataset has been divided into 800 high resolution images for training, 100 high resolution images for validation and 100 high resolution images for testing. For conducting the experiments for this paper we had resized these images to a size of 256x256 for the high resolution images. For the corresponding low resolution images we resize the 256x256 images to half their size and then rescale to 256x256 i.e 50 percent of the original 256x256 HR image. A sample subset of images can be seen in Figure \ref{div2kset}.

\begin{figure}[htbp]
\centerline{\includegraphics[width=8cm, height=4cm]{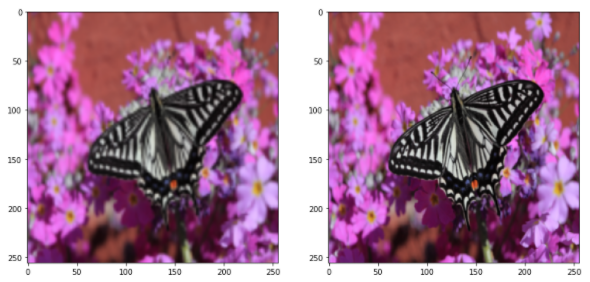}}
\caption{(Left) 256x256 Rescaled LR (Scale : 50) 
(Right) 256x256 Original HR Resized.}
\label{inputsample}
\end{figure}
For the training set we had taken the original DIV2K training set which consists of 800 2K
resolution images. For computation purposes we had resized each of these images to 256x256
resolution. The LR input for these corresponding HR images were generated by rescaling the HR images by a factor of 50 and then resized back to 256x256 size. For faster computation, we had to create a generator function which randomly selects a batch of 10 images from the dataset and converts it to the required input format. A sample input LR and HR pair can be seen in Figure \ref{inputsample}.

\subsection{Model Architecture}
\begin{figure}[htbp]
\centerline{\includegraphics[width=6cm, height=11cm]{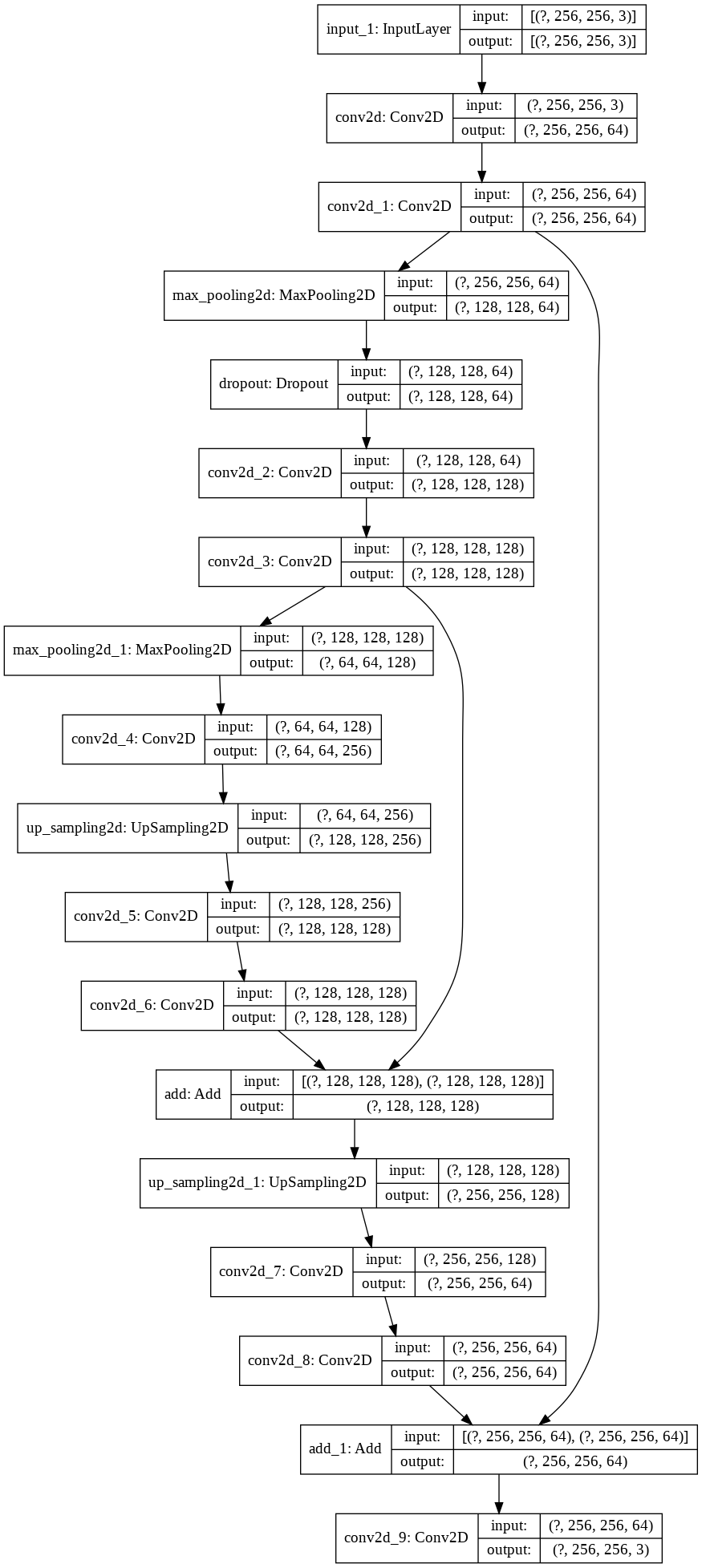}}
\caption{Model Architecture of Vanilla CNN U-Net.}
\label{unetmod}
\end{figure}

\begin{figure}[htbp]
\centerline{\includegraphics[width=8cm, height=4cm]{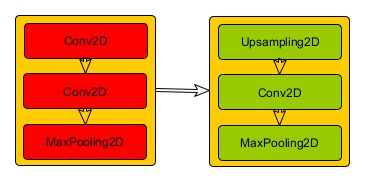}}
\caption{Left: Downscaling Block Right: Upscaling Block.}
\label{unet_block}
\end{figure}

For the purpose of Super Resolution we had first trained a vanilla CNN U-Net Model as shown in figure \ref{unetmod}. The U-Net model consists of two convolutional downscaling blocks followed by two upscaling blocks. The output of each downscaling block is added to the corresponding upscaling block. We add a dropout layer to the first downscaling block to reduce computation. The downscaling and Upscaling blocks are shown in figure \ref{unet_block}. A pretrained MR-CNN was used for obtaining the segmentation masks. The output of the MR-CNN on the LR Image is preprocessed to obtain the segmented image which is then used as input for the U-Net model. A pretrained implementation of Mask R-CNN \cite{matter} was used for the purpose of foreground segmentation.

\pagebreak
\subsection{Loss Function and Optimizer}
The MSE Loss or Mean Squared Error Loss is used for calculating pixel-wise loss between the original HR image and the U-Net output. This is calculated and compared across all three channels in an RGB image. The mean of each pixel difference is calculated and then squared.
\\
\\
\begin{gather*} 
\mathit{MSE}=\frac{1}{mn} \sum_{i=1}^m \sum_{j=1}^n (x_{ij}-y_{ij})^2 \\[2ex]
\end{gather*}
 
The optimizer used here is the Adam Optimizer. The advantage of using the Adam optimizer is that the optimizer has adaptive learning rates for each parameter. The optimizer estimates moments (moving average of the parametes)and uses them to optimize a function. It is more robust since it is a combination of RMSprop and SGD with momentum Optimizers. Moreover it is beneficial for models with large number of parameters or for training on large datasets.

\subsection{Training}
\begin{figure}[htbp]
\centerline{\includegraphics[width=8cm, height=6cm]{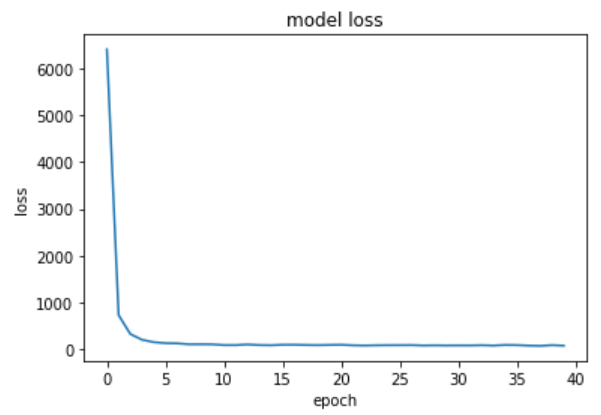}}
\caption{Training Loss Curve}
\label{losscurv}
\end{figure}

We trained the vanilla U-Net model for 40 epochs with a step size of 20. The model quickly dropped from a loss of 6402.0 to a final loss of 83.72 as shown in figure \ref{losscurv}. Further training can decrease the loss value. The loss value per epoch for every 4 epochs is shown in table \ref{losstab}. To prevent the model from stagnating at a particular loss value we had leveraged Tensorflow's ReduceLRonPlateau functionality to modify the learning rate if the model starts to become stationary at a particular loss value/range. 
\begin{center}
\begin{table}[h]
\label{losstab}
\fontsize{9}{11}\selectfont
\begin{tabular}{ | m{4cm} | m{4cm}|  } 
  \hline
   \textbf{Epoch} &\textbf{Loss} \\ 
  \hline
 1 & 6402.04 \\ 
  \hline
 5 & 158.56 \\ 
  \hline
 9 & 114.28 \\ 
  \hline
 13 & 109.07 \\ 
  \hline
 17 & 103.53 \\ 
  \hline
 21 & 103.12 \\ 
  \hline
 25 & 94.47 \\ 
  \hline
 29 & 91.10 \\ 
  \hline
 33 & 93.99 \\ 
  \hline
 37 & 85.24 \\ 
  \hline
\textbf{40} & \textbf{83.72} \\ 
  \hline
\end{tabular}
\\
\\
\\
\caption{Training Loss for every 4 epochs and final epoch loss.}
\end{table}
\end{center}

\section{Evaluation Metrics}
\subsection{PSNR - Peak Signal-to-Noise Ratio}
Peak signal-to-noise ratio is a concept that shows the ratio between the power of a signal and the noise that affects its representation. In terms of images, it is the ratio between the largest possible power of an image and the largest power of corrupting noise (image noise such as jittered edges, deformities, etc.) that affects the quality and appearance of an image.
\begin{equation} 
\mathit{PSNR}(x,y)=\frac{10\log_{10}[\max(\max(x),\max(y))]^2}{\abs{x-y}^2}
\end{equation}
\\
\subsection{SSIM - Structural Similarity Index}
The Structural Similarity Index (SSIM) \cite{ssim} is a method of measuring the similarity between images. It can be used to evaluate the quality of one image over another. It is an image quality metric that analysis the significant impact of three major characteristics present in an image: luminance, contrast and structure.
\begin{equation}
  \SSIM(x,y) = \frac{(2\mu_x\mu_y + C_1) + (2 \sigma _{xy} + C_2)} 
    {(\mu_x^2 + \mu_y^2+C_1) (\sigma_x^2 + \sigma_y^2+C_2)}
\end{equation}
\\
\subsection{Universal Image Quality Index}
The UQI metric \cite{uqi} is computed by considering the effects of image distortion on the quality of subjective measurements. Experiments have revealed that it performs significantly better compared to the commonly used mean squared error metric.
\begin{equation}
  UQI(x,y) = \frac{(4\sigma_(xy))\hat{x}\hat{y}} 
    {(\sigma_x^2 + \sigma_y^2) (\hat{x}^2 + \hat{y}^2)}
\end{equation}
\\
\section{Results and Discussion}
\begin{figure}[htbp]
\centerline{\includegraphics[width=9cm, height=5cm]{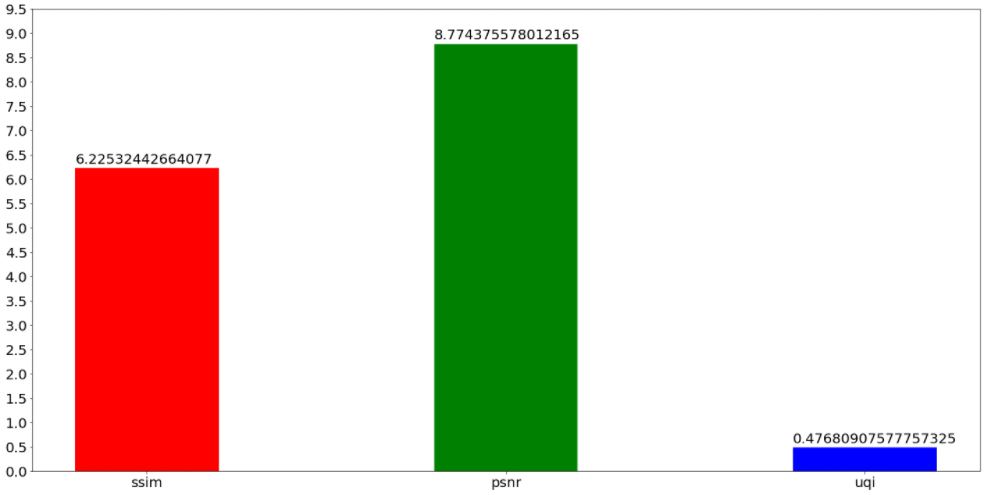}}
\caption{Evalutation Metric Increase after Super Resolution on 50 percent Quality Images}
\label{metgraph}
\end{figure}

\begin{figure}[htbp]
\centerline{\includegraphics[width=9cm, height=3cm]{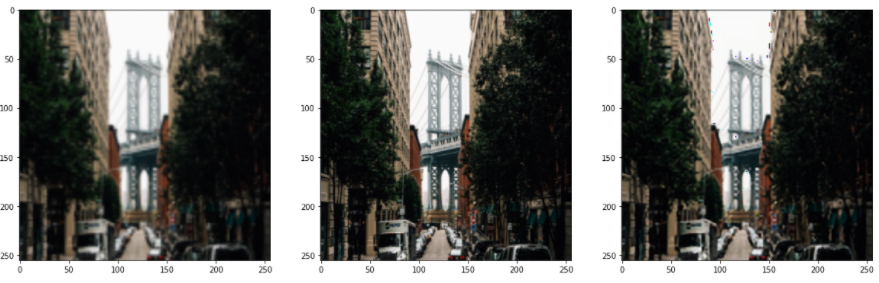}}
\caption{Left-Right : Original LR Image (0.8152 SSIM|25.5518 PSNR), Original HR Image, Predicted SR Image (0.8724 SSIM|26.2194 PSNR)}
\label{ex1}
\end{figure}

\begin{figure}[htbp]
\centerline{\includegraphics[width=9cm, height=3cm]{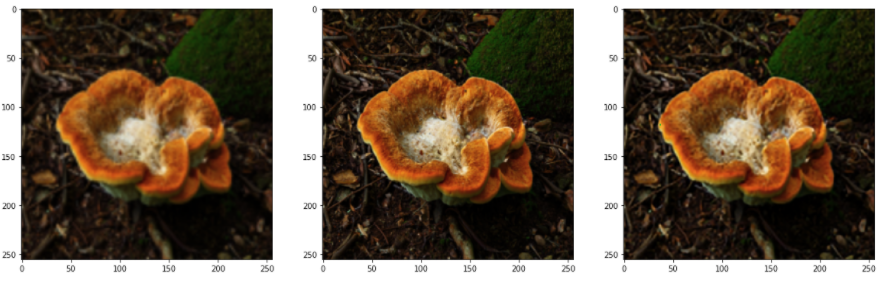}}
\caption{Left-Right : Original LR Image (0.8499 SSIM|29.0403 PSNR), Original HR Image, Predicted SR Image (0.9099 SSIM|31.2586 PSNR)}
\label{ex2}
\end{figure}

The U-Net model was evaluated on the validation set which consists of 100 HR Images and their corresponding computed LR images as input. After the inference the mean increase in SSIM of the validation set was calculated. We had achieved an average increase of about 6.2253 in the SSIM score and an increase of about 8.7743 in the PSNR score and an increase of about 0.4768 in UQI metrics. The overall increase for each metric is shown in graph figure \ref{metgraph}. The algorithm for metric percentage increase is shown in algorithm \ref{metricval}. Some samples of the original SR Output is shown in Figures \ref{ex1} and \ref{ex2}. 
\linebreak
\begin{algorithm}
\KwData{$Validation Image Batch (X,Y)$}
\KwResult{$Average Increase Metric Score$}
$X \gets x$\;
$Y \gets y$\;
$N \gets len(Batch)$\;
$overallpercent = 0$
\For{(x,y) in (X,Y)}{
$prediction = Model(x)$
$original score = Metric(x,y)$
$score = Metric(prediction,y)$
\\
$scoreincrease = (score - originalscore)/originalscore *100$
$overallpercent+=scoreincrease$
}
$overallpercent=overallpercent/N$
\linebreak
\caption{Validation Method for a particular Metric}\label{metricval}
\end{algorithm}

\begin{figure}[h]
\centerline{\includegraphics[width=9cm, height=3cm]{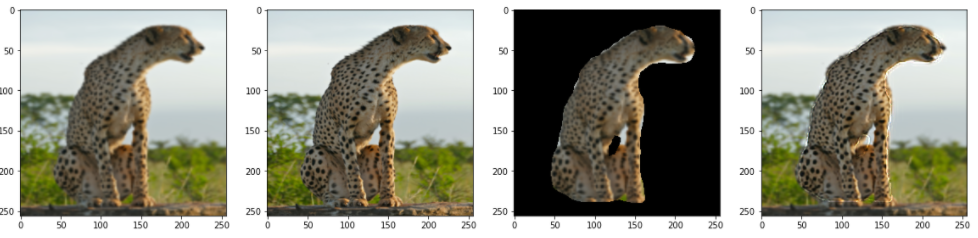}}
\caption{Left-Right : Original LR Image, Original HR Image, Masked LR Image, Masked SR + Background LR\\
Below: Metric Values of Original LR and Foreground SR}
\label{fr1}
\end{figure}

\begin{figure}[h]
\centerline{\includegraphics[width=9cm, height=3cm]{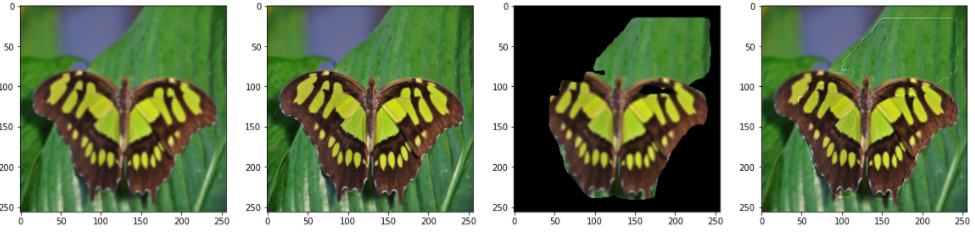}}
\caption{Left-Right : Original LR Image, Original HR Image, Masked LR Image, Masked SR + Background LR\\
Below: Metric Values of Original LR and Foreground SR}
\label{fr2}
\end{figure}

\begin{figure}[htbp]
\centerline{\includegraphics[width=8cm, height=8cm]{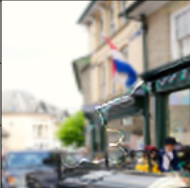}}
\caption{Jittered Edge between Background LR and Masked SR Region}
\label{edgejit}
\end{figure}

We then compared the metrics between the vanilla CNN U-Net for Super Resolution and our Forground Super Resolution Flow. The LR Images are passed into the MR-CNN and the Masked regions are passed into the U-Net model and we evaluate the metrics on the combined forground segmented SR output and LR background image. In some cases the output showed an increase in quality. In other images the quality decreased drastically. Since the focus of our paper is just to increase foreground or object in context resolution this can be ignored. Here are few examples with SSIM and PSNR scores and examples with all three metrics before and after foregorund super resolution in figures \ref{fr1} and \ref{fr2}.
\begin{figure}[htbp]
\centerline{\includegraphics[width=9cm, height=3cm]{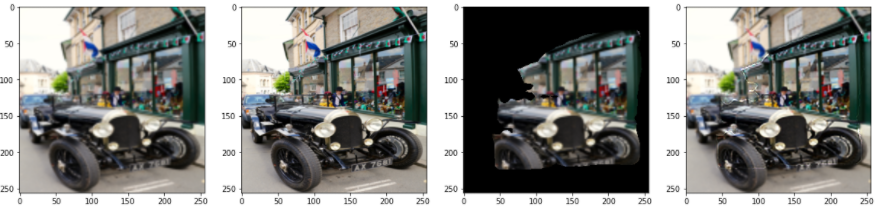}}
\caption{Left-Right : Original LR Image (0.8716 SSIM|25.1606 PSNR), Original HR Image, Masked LR Image, Masked SR + Background LR (0.8919 SSIM|25.6003 PSNR)}
\label{fig}
\end{figure}
\begin{figure}[h]
\centerline{\includegraphics[width=9cm, height=3cm]{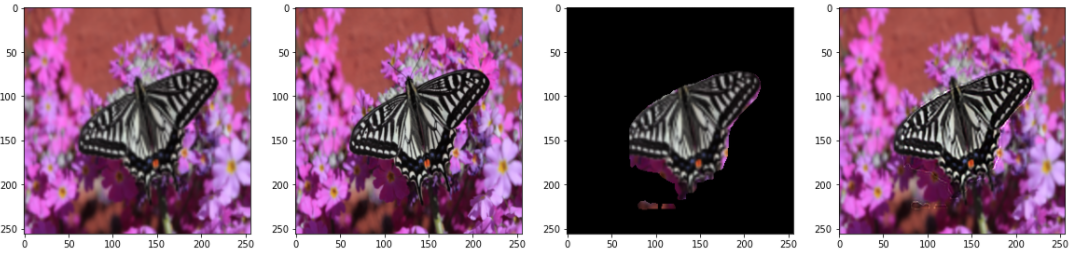}}
\caption{Left-Right : Original LR Image (0.9306 SSIM|26.1428 PSNR), Original HR Image, Masked LR Image, Masked SR + Background LR (0.9382 SSIM|27.2623 PSNR)}
\label{fig}
\end{figure}
\begin{center}
\begin{tabular}{ | m{1.5cm} | m{1.5cm}| m{1.5cm} |m{1.5cm} | } 
  \hline
   \textbf{Image} &\textbf{SSIM} & \textbf{PSNR} & \textbf{UQI} \\ 
  \hline
 LR and HR & 0.9252 & 29.9149 & 0.9945 \\ 
  \hline
LR and Output & 0.9258 & 27.4436 & 0.9935 \\ 
  \hline
\end{tabular}
\end{center}
\begin{center}
\begin{tabular}{ | m{1.5cm} | m{1.5cm}| m{1.5cm} |m{1.5cm} | } 
  \hline
   \textbf{Image} &\textbf{SSIM} & \textbf{PSNR} & \textbf{UQI} \\ 
  \hline
 LR and HR & 0.9056 & 29.2197 & 0.9932 \\ 
  \hline
LR and Output & 0.8888 & 29.2082 & 0.9923 \\ 
  \hline
\end{tabular}
\end{center}
 But in some cases the imposed super resolution mask with the original LR creates unwanted white and jittery boundary edges as shown in Figure \ref{edgejit}. This might be the main cause decrease of quality. This can be further solved by applying smoothing filters to merge those edges with the background.
\\
\\
\section{Conclusion and Future Work}
Using the DIV2K dataset, the application of Foreground Image Super Resolution was explored using a Vanilla CNN U-Net as the SR Model and MR-CNN for Foreground Segmentation. The U-Net trained was able to increase the quality of the LR Images easily. For the foreground or portrait images the metric values showed a slight increase in SSIM value cause the model specifically increases only the quality of the object in context (segmented object). This was possible cause even if the background of the image is increased in quality, Due to blurred effect in portrait there is not much difference between the background of a LR image and a HR image. Thus the metric increase for the background does not matter. But the PSNR metric showed a drastic decrease due to the presence of jittered white edges (borders) when super imposing the SR Segmented Image back to the LR image. These edges add to the noise factor of the image and thus we see a decrease in PSNR values.
Further developments can be made to negate these blurred edges such as using filtered blurring or merging of edges with the neighboring pixels or the background. Moreover we have only implemented a simple U-Net architecture for our experiments due to constrained CPU and GPU resources. By implementing state-of-the-art architectures such as WDSR, EDSR, etc. we can achieve 2K-4K resolution quality for the foreground images. These improvements combined with the base architecture can be applied to various applications such as custom creation of portrait images, superimposing of images on to various backgrounds similar to the video filters provided by video conferencing platforms such as Zoom, Microsoft Meetings and other social media platforms such as Snapchat, Instagram, etc.
\pagebreak
\section*{Acknowledgements}
We thank Ydealogy Ventures for providing us with the required computational resources. The company's resources as well as their constant feedback and 
insights while working on this paper were highly valuable.


\begin{thebibliography}{99} 
\bibitem{SRCNN} 
Chao Dong, Chen Change Loy, Kaiming He, Xiaoou Tang
\newline
\textit{Image Super-Resolution Using Deep Convolutional Networks}. 
CoRR, 2015.

\bibitem{cnn} 
Alex Krizhevsky, Ilya Sutskever, and Geoffrey Hinton
\newline
\textit{ImageNet Classification with Deep Convolutional Networks}. 
ILSVRC, 2012.

\bibitem{vdsr} 
J. Kim, J. Kwon Lee, and K. Mu Lee
\newline
\textit{Accurate image super-resolution
using very deep convolutional networks}. 
Proceedings of the IEEE
Conference on Computer Vision and Pattern Recognition, 2016

\bibitem{fsrcnn} 
C. Dong, C. C. Loy, and X. Tang
\newline
\textit{Accelerating the super-resolution
convolutional neural network}. 
Proceedings of the European Conference
on Computer Vision, 2016

\bibitem{espcn} 
Wenzhe Shi, Jose Caballero, Ferenc Huszár, Johannes Totz, Andrew P. Aitken, Rob Bishop, Daniel Rueckert, Zehan Wang
\newline
\textit{Real-Time Single Image and Video Super-Resolution Using an Efficient Sub-Pixel Convolutional Neural Network}. 
CoRR, 2016

\bibitem{div2k} 
Agustsson, Eirikur and Timofte, Radu
\newline
\textit{NTIRE 2017 Challenge on Single Image Super-Resolution: Dataset and Study}. 
The IEEE Conference on Computer Vision and Pattern Recognition (CVPR) Workshops, 2017.

\bibitem{unet} 
Olaf Ronneberger, Philipp Fischer, Thomas Brox
\newline
\textit{U-Net: Convolutional Networks for Biomedical Image Segmentation}. 
CoRR, 2015.

\bibitem{mrcnn} 
Kaiming He, Georgia Gkioxari, Piotr Doll, Ross B. Girshick
\newline
\textit{Mask {R-CNN}}. 
CoRR, 2018.

\bibitem{resnet} 
Kaiming He, Xiangyu Zhang, Shaoqing Ren, Jian Sun
\newline
\textit{Deep Residual Learning for Image Recognition}. 
CoRR, 2015.

\bibitem{edsr} 
Bee Lim, Sanghyun Son, Heewon Kim, Seungjun Nah, Kyoung Mu Lee
\newline
\textit{Enhanced Deep Residual Networks for Single Image Super-Resolution}. 
CoRR, 2015.


\bibitem{wdsr} 
Jiahui Yu, Yuchen Fan, Jianchao Yang, Ning Xu, Zhaowen Wang, Xinchao Wang, Thomas Huang
\newline
\textit{Wide Activation for Efficient and Accurate Image Super-Resolution}. 
CoRR, 2018.

\bibitem{rhbm} 
Shanchuan Lin, Andrey Ryabtsev, Soumyadip Sengupta, Brian Curless, Steve Seitz, Ira Kemelmacher-Shlizerman
\newline
\textit{Real-Time High-Resolution Background Matting}. 
CoRR, 2020.

\bibitem{deeplab} 
Liang-Chieh Chen, George Papandreou, Florian Schroff, Hartwig Adam
\newline
\textit{Rethinking Atrous Convolution for Semantic Image Segmentation}. 
CoRR, 2017.

\bibitem{ssim} 
Zhou Wang, A. C. Bovik, H. R. Sheikh, E. P. Simoncelli
\newline
\textit{Image quality assessment: from error visibility to structural similarity}. 
IEEE Transactions on Image Processing, 2004.

 \bibitem{uqi} 
Zhou Wang, A.C. Bovik
\newline
\textit{A universal image quality index}. 
IEEE Signal Processing Letters, 2002

 \bibitem{matter} 
Waleed Abdulla
\newline
\textit{Mask R-CNN for object detection and instance segmentation on Keras and TensorFlow}. 
Github, 2017

 \bibitem{vggfor} 
Ajmal Shahbaz, Kang-Hyun Jo
\newline
\textit{Deep Foreground Segmentation using Convolutional Neural Network}. 
IEEE 28th International Symposium, 2019

 \bibitem{scenerem} 
Marc Braham, Marc Van Droogenbroeck
\newline
\textit{Deep background subtraction with scene-specific convolutional neural networks}. 
IWSSIP, 2016

\end{thebibliography}
\end{document}